\documentclass[sigconf]{acmart}
\AtBeginDocument{%
  }

\copyrightyear{2025}
\acmYear{2025}
\setcopyright{acmlicensed}
\acmConference[MM '25] {Proceedings of the 33rd ACM International Conference on Multimedia}{October 27--31, 2025}{Dublin, Ireland.}
\acmBooktitle{Proceedings of the 33rd ACM International Conference on Multimedia (MM '25), October 27--31, 2025, Dublin, Ireland}
\acmISBN{979-8-4007-2035-2/2025/10}
\acmDOI{10.1145/XXXXXX.XXXXXX}

\acmSubmissionID{4558}

\usepackage{booktabs}
\usepackage{multirow}
\usepackage{makecell}
\usepackage{array}
\usepackage{colortbl}
\usepackage{xcolor}
\usepackage{graphicx}
\usepackage{subcaption}
\usepackage{enumitem}

\begin{document}

\title{HiProbe-VAD: Video Anomaly Detection via Hidden States Probing in Tuning-Free Multimodal LLMs}

\author{Zhaolin Cai}
\orcid{0009-0001-9837-577X}
\affiliation{
  \institution{Xinjiang University}
  \city{Urumqi}
  \state{Xinjiang}
  \country{China}
}
\email{107552301311@stu.xju.edu.cn}
\author{Fan Li}
\orcid{0000-0002-7566-1634}
\affiliation{
  \institution{Xi'an Jiaotong University}
  \city{Xi'an}
  \state{Shaanxi}
  \country{China}
}
\email{lifan@mail.xjtu.edu.cn}
\author{Ziwei Zheng}
\orcid{0009-0000-4896-3293}
\affiliation{
  \institution{Xi'an Jiaotong University}
  \city{Xi'an}
  \state{Shaanxi}
  \country{China}
}
\email{ziwei.zheng@stu.xjtu.edu.cn}
\author{Yanjun Qin}
\authornote{*Corresponding author}
\orcid{0000-0001-5011-8697}
\affiliation{
  \institution{Xinjiang University}
  \city{Urumqi}
  \state{Xinjiang}
  \country{China}
}
\email{qinyanjun@xju.edu.cn}

\renewcommand{\shortauthors}{Cai Z et al.}

\begin{abstract}
Video Anomaly Detection (VAD) aims to identify and locate deviations from normal patterns in video sequences. Traditional methods often struggle with substantial computational demands and a reliance on extensive labeled datasets, thereby restricting their practical applicability. To address these constraints, we propose HiProbe-VAD, a novel framework that leverages pre-trained Multimodal Large Language Models (MLLMs) for VAD without requiring fine-tuning. In this paper, we discover that the intermediate hidden states of MLLMs contain information-rich representations, exhibiting higher sensitivity and linear separability for anomalies compared to the output layer. To capitalize on this, we propose a Dynamic Layer Saliency Probing (DLSP) mechanism that intelligently identifies and extracts the most informative hidden states from the optimal intermediate layer during the MLLMs reasoning. Then a lightweight anomaly scorer and temporal localization module efficiently detects anomalies using these extracted hidden states and finally generate explanations. Experiments on the UCF-Crime and XD-Violence datasets demonstrate that HiProbe-VAD outperforms existing training-free and most traditional approaches. Furthermore, our framework exhibits remarkable cross-model generalization capabilities in different MLLMs without any tuning, unlocking the potential of pre-trained MLLMs for video anomaly detection and paving the way for more practical and scalable solutions.
\end{abstract}



\maketitle

\begin{figure}[t!]
    \centering
    \includegraphics[width=\linewidth]{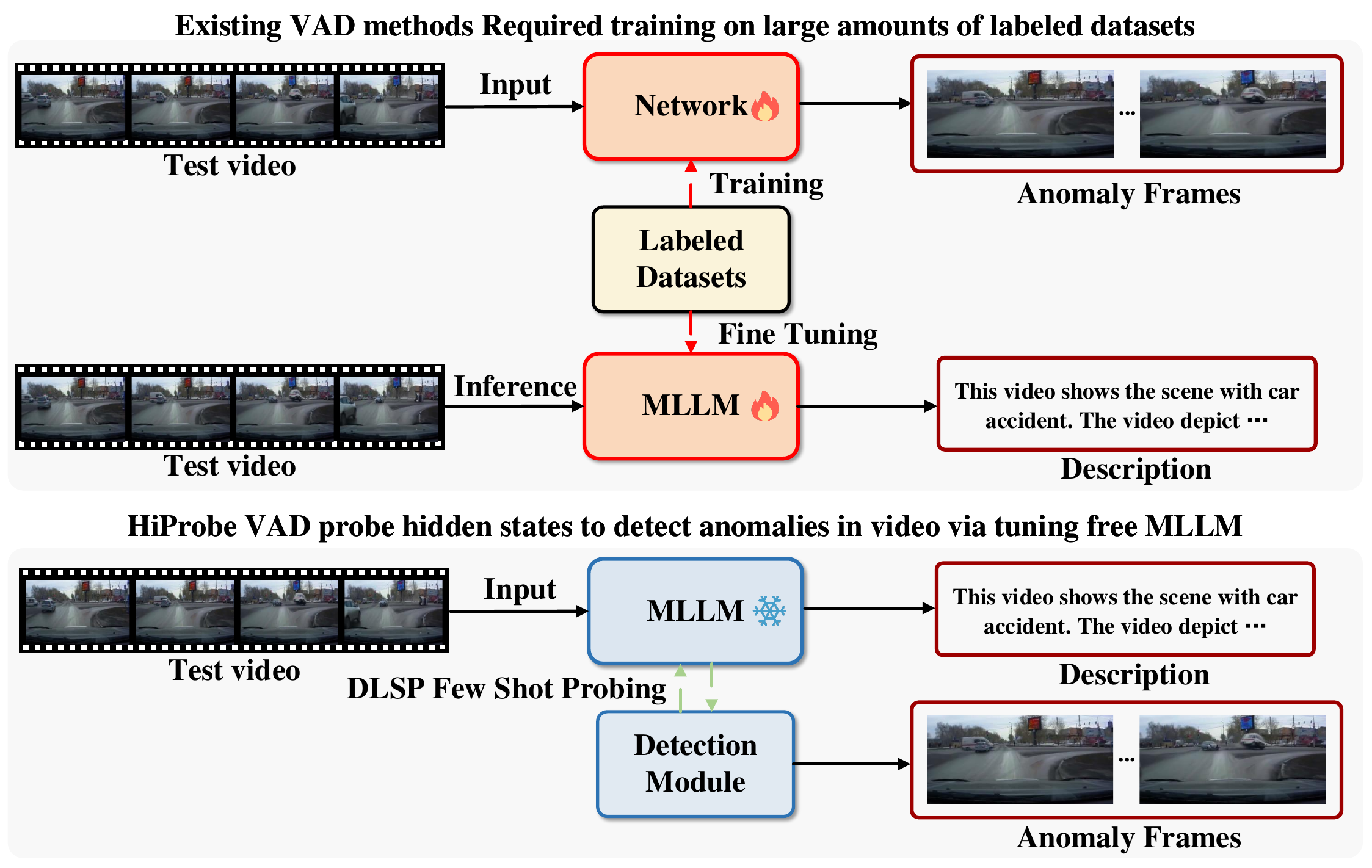}
    \caption{HiProbe-VAD utilizes hidden states in the intermediate layer of MLLMs to efficiently detect anomalies in videos.}
    \Description{A diagram comparing HiProbe-VAD to other video anomaly detection methods. HiProbe-VAD utilizes hidden states in the intermediate layer of MLLMs to efficiently detect anomalies in videos.}
    \label{fig:comparemethods}
\end{figure}

\section{Introduction}
Video Anomaly Detection (VAD) aims to locate events or behaviors in videos that deviate from normal patterns, which is crucial for applications spanning video surveillance \cite{sultani2018RealWorld}, industrial quality inspection \cite{roth2022Total}, and autonomous driving \cite{yao2023DoTA, bogdoll2022Anomaly}. While achieving high accuracy is essential, inherent complexity and dataset-dependent nature of anomalies pose significant challenges to VAD systems. Existing deep learning-based VAD approaches encompass supervised, weakly supervised, and unsupervised learning paradigms. Supervised methods \cite{liu2018Future,landi2019Anomaly} achieve high accuracy but require extensive and costly frame-level annotations. Weakly supervised methods \cite{feng2021MIST,lv2023Unbiased, joo2023CLIPTSA} mitigate this labeling burden by leveraging limited or video-level labels, often at the expense of detection granularity or performance. Unsupervised methods \cite{liu2021Hybrid, lv2021Learning, tur2023Exploring} learn normal patterns from unlabeled data to detect anomalies. These methods can struggle with anomalies but require substantial labeled data for pre-training, potentially limiting their deployment(see Fig. \ref{fig:comparemethods}). These limitations highlight the ongoing need for VAD solutions with reduced data dependency and improved efficiency.

The recent emergence of Multimodal Large Language Models (MLLMs) \cite{zhu2023MiniGPT4, liu2023Visual, wu2024DeepSeekVL2, xu2025Qwen25omni} has presented novel avenues for various vision tasks due to their remarkable ability to jointly process and reason about visual and textual information, offering promising new directions for VAD \cite{chen2024How, zheng2025spotrisksspeakingunraveling}. Prior works have explored adapting these models via fine-tuning or prompt engineering for specific anomaly detection tasks \cite{zhang2024HolmesVAD, zhang2024HolmesVAU, pu2024Learning}. However, these approaches typically suffer from two main drawbacks: (1) the need for task-specific fine-tuning on VAD datasets, which is computationally expensive and often requires substantial labeled datasets. (2) an over-reliance on text representations derived from visual inputs, potentially leading to a loss of critical visual details during the inference and resulting in incomplete or biased video understanding. 

Recent studies in the field of Natural Language Processing have revealed that intermediate layers of Large Language Models often contain richer and more transferable representations compared to output layers \cite{fan2024Not, chen2024bigger, park2023Linear}. These intermediate layers have shown superior performance across various tasks \cite{skean2025Layer, skean2024Does}, suggesting they capture a more nuanced understanding of the input data \cite{orgad2024LLMs, alain2018Understanding}. Inspired by these findings in LLMs, we hypothesize that the intermediate hidden states within MLLMs similarly encapsulate rich information, potentially even more effective for discerning video anomalies compared to the final output layers. We further posit that this richer information within the intermediate layers of pre-trained MLLMs might inherently contain or better activate the model's pre-existing capacity for distinguishing between normal and anomalous visual patterns, even without explicit fine-tuning for video anomaly detection. This potential to leverage the inherent anomaly detection capabilities through intermediate representations lays a crucial foundation for exploring a novel tuning-free framework via MLLMs for video anomaly detection. 

In this paper, we present a systematic analysis of the intermediate information within MLLMs and reveal a key finding: intermediate hidden states within MLLMs exhibit improved sensitivity and linear separability to anomalies compared to output layers. We therefore define this observation as Intermediate Layer Information-rich Phenomenon. Based on this finding, we propose \textbf{Hi}dden-state \textbf{Prob}ing framework for \textbf{V}ideo \textbf{A}nomaly \textbf{D}etection (HiProbe-VAD), a tuning-free framework that harnesses pre-trained MLLMs for VAD. HiProbe-VAD employs a Dynamic Layer Saliency Probing (DLSP) module to extract hidden states from the intermediate layers and dynamically select the most effective layer during a single forward pass of the MLLM. Subsequently, a lightweight anomaly scorer based on logistic regression and temporal localization module are integrated to deliver efficient detection and precise localization. Finally, to provide interpretable insights into the detected anomalies, anomaly frames and normal frames are input to an auto-regression process to generate detailed textual descriptions of the detected events. We evaluate the effectiveness of our framework through extensive experiments on UCF-Crime \cite{sultani2018RealWorld} and XD-Violence \cite{wu2020Not} datasets. These datasets cover diverse real-world scenarios, providing a robust testbed for evaluating VAD performance. Through comprehensive experiments, we demonstrate the effectiveness of HiProbe-VAD framework in video anomaly detection.

Our main contributions are as follows:

\begin{itemize}[noitemsep, topsep=0.5em]
    \item We present the first systematic quantification of the 'intermediate layer information-rich phenomenon' in MLLMs for video anomaly detection, demonstrating intermediate hidden states outperform output layers in anomaly sensitivity and separability, challenging the inherent limitations of output-layer dependent MLLM approaches.
    \item We propose HiProbe-VAD, a novel tuning-free VAD framework that effectively leverages the intermediate information within pre-trained MLLMs, enabling anomaly detection without fine-tuning the MLLM, while requiring only minimal coarse labeled data to train a lightweight anomaly scorer.
    \item Experiments show that HiProbe-VAD achieves competitive results compared to state-of-the-art tuning-free, unsupervised, and self-supervised VAD methods. Our framework exhibits strong cross-model generalization capabilities by demonstrating its robustness and adaptability across various MLLM architectures.
\end{itemize}

\section{Related Works}

\subsection{Traditional Video Anomaly Detection}
VAD is the task of identifying deviated frames from normal patterns in video \cite{lu2013Abnormal, mehran2009Abnormal, jiao2023Survey}, which is a task extensively studied in multimedia research. Existing VAD methods can be classified into supervised, weakly supervised, and unsupervised. Supervised methods \cite{liu2018Future,landi2019Anomaly} achieve high accuracy through detailed frame-level annotations but face significant limitations due to the prohibitive annotation costs. Weakly supervised methods \cite{lv2023Unbiased, li2022Scaleaware, wang2024Weakly, zhang2024GlanceVAD} leverage video-level labels to train and detect abnormal videos but struggle with subtle anomalies and may exhibit biased predictions. Unsupervised approaches \cite{tur2023Unsupervised}, like one-class learning \cite{hasan2016Learning, xu2017Detecting, yang2023Video}, train solely on normal data and flag deviations during testing; despite their flexibility, these methods often yield high false positives due to the challenge of completely modeling normal variability. 

\subsection{Video Anomaly Detection based on LLMs and MLLMs}
The recent emergence of Large Language Models (LLMs) \cite{touvron2023llama, vaswani2017attention, bubeck2023sparks} and Multimodal Large Language Models (MLLMs) \cite{zhu2023MiniGPT4, liu2023Visual, li2023BLIP2} has introduced novel directions and approaches for video anomaly detection \cite{zhang2023VideoLLaMA, wu2024Openvocabulary}. Most approaches fine-tune pre-trained MLLMs to perform anomaly detection and analysis \cite{lv2024Video, zhang2024HolmesVAD, zhang2024HolmesVAU, yuan2024Surveillance}, requiring substantial labeled data and computational resources. Some methods like \cite{zanella2024Harnessing} try to explore tuning-free method with generated textual descriptions from video frames with VLM and infer anomalies based on these descriptions with LLM, \cite{ye2024VERA} also try to guide pre-trained VLM to reason better via verbalized learning, but the reliance on text probably lead to overlooking subtle visual cues. While promising, these methods often remain limited by their dependence on either text outputs or the need for fine-tuning, thus underutilizing the full multimodal potential of MLLMs.

\subsection{Analysis of Intermediate Layers in LLMs}
Recent studies on the intermediate layers of Large Language Models (LLMs) \cite{jin2024Exploring, ju2024How, katz2023VISIT, merullo2024Talking}, revealing that they often contain richer and more informative representations compared to the final layers \cite{orgad2024LLMs, sun2024Massive, yang2025Exploring}. Research has shown that intermediate layers outperform final layers across diverse applications, potentially due to their ability to balance information retention and noise reduction through mechanisms like compression and feature distillation \cite{skean2025Layer, skean2024Does, chen2024bigger}. Furthermore, intermediate layers have been found to play a crucial role in complex reasoning tasks, as models trained for such purposes tend to preserve more contextual information at these depths, thereby enhancing multi-step inference capabilities \cite{park2023Linear, orgad2024LLMs, zhang2024Investigating}. Building on these insights from general LLMs, we hypothesize that the intermediate hidden states of pre-trained MLLMs also contain richer and more informative representations, which motivates our exploration of a novel tuning-free framework for VAD by effectively probing these intermediate layers.

\section{Information-Rich Phenomenon for Video Anomaly Detection}

Motivated by the demonstrated richness of intermediate layers in Large Language Models, we investigate whether a similar phenomenon exists within Multimodal Large Language Models (MLLMs) for video anomaly detection. We hypothesize that these intermediate layers might offer a more direct and nuanced representation of anomalies compared to the final output layers, which are typically optimized primarily for text generation, potentially leading to improved anomaly detection performance.

\subsection{Exploring Intermediate Layer Representations in MLLMs}

To validate our hypothesis in MLLMs for video anomaly detection, we conduct systematic exploration of the hidden state representations extracted from different layers within pre-trained MLLMs. For each input video $V$ from benchmark VAD datasets (XD-Violence \cite{wu2020Not} and UCF-Crime \cite{sultani2018RealWorld}), we perform a single forward pass using the pre-trained MLLM (InternVL2.5 \cite{chen2024Intern}). During this pass, we extract hidden states $\mathbf{h}_l$ from each layer $l$. We then evaluate the effectiveness of these representations in distinguishing between normal and anomalous videos using statistical and geometric analyses, focusing on different aspects of feature quality. The following subsections detail our methodologies and metrics.

\subsubsection{Statistical Quantification for VAD}

To quantify the information captured across layers statistically, we focused on quantifying key properties of the extracted hidden states $\mathbf{h}_l$. We employed the following metrics, chosen to capture different aspects of feature quality for anomaly detection:

\begin{figure}[t!]
    \centering
    \includegraphics[width=\linewidth]{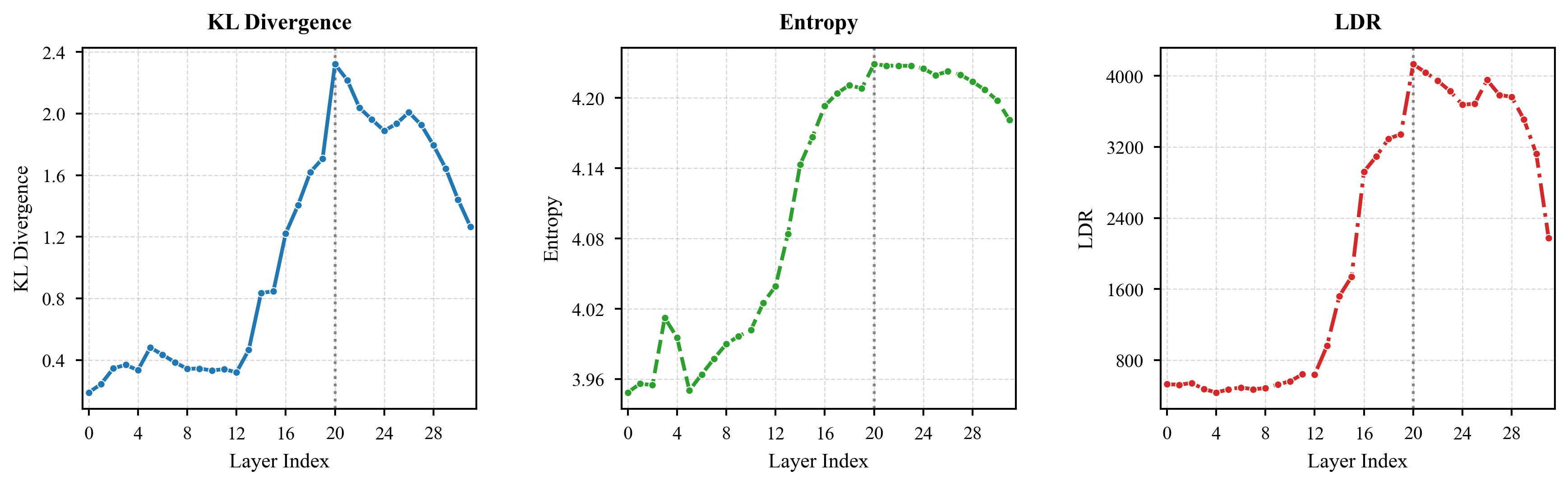}
    \caption{Analysis of hidden state properties across layers of a pre-trained MLLM (InternVL2.5) on the XD-Violence dataset. Kullback-Leibler (KL) Divergence, Local Discriminant Ratio (LDR), and Entropy metrics consistently exhibit distinct patterns peaking around intermediate layer 20.}
    \label{fig:threemetrics}
    \Description{Analysis of hidden state properties across layers of a pre-trained MLLM (InternVL2.5) on the XD-Violence dataset. Kullback-Leibler (KL) Divergence, Local Discriminant Ratio (LDR), and Entropy metrics consistently exhibit distinct patterns peaking around intermediate layer 20.}
\end{figure}

\begin{itemize}[noitemsep,leftmargin=*, topsep=0.5em]
    \label{metricss}
    \item \textbf{Anomaly Sensitivity via KL Divergence:} The Kullback-Leibler (KL) divergence quantifies the statistical distinguishability between normal and anomalous features. For each feature dimension $d$ at layer $l$, we assume that the hidden states of normal samples ($\mathbf{h}_l^N$) and anomalous samples ($\mathbf{h}_l^A$) are approximately Gaussian distributed, i.e., $\mathcal{N}(\mu_{l,d}^N, (\sigma_{l,d}^N)^2)$ and $\mathcal{N}(\mu_{l,d}^A, (\sigma_{l,d}^A)^2)$, respectively. The KL divergence between these two Gaussian distributions for the $d$-th dimension is given by:
    \begin{equation}
         D_{\text{KL}}^{(d)}(l) = \frac{1}{2} \left[ \log \left( \frac{(\sigma_{l,d}^A)^2}{(\sigma_{l,d}^N)^2} \right) + \frac{(\sigma_{l,d}^N)^2 + (\mu_{l,d}^N - \mu_{l,d}^A)^2}{(\sigma_{l,d}^A)^2} - 1 \right]. 
    \end{equation}

    The overall anomaly sensitivity for layer $l$ is then the average KL divergence across all feature dimensions $D$:
    \begin{equation}
        D_{\text{KL}}(l) = \frac{1}{D} \sum_{d=1}^{D} D_{\text{KL}}^{(d)}(l).
    \end{equation}
    A higher $D_{\text{KL}}(l)$ indicates a greater distributional difference between normal and anomalous features at layer $l$.

    \item \textbf{Class Separability via Local Discriminant Ratio:} The Local Discriminant Ratio (LDR) measures the ability of features to linearly separate different classes. For each feature dimension $d$ at layer $l$, we calculate a LDR as the ratio of the squared difference between the means of normal ($\mu_{l,d}^N$) and anomalous ($\mu_{l,d}^A$) features to the sum of their variances $((\sigma_{l,d}^N)^2$ and $(\sigma_{l,d}^A)^2)$, with a small constant $\epsilon$ added for numerical stability:
    \begin{equation}
        \text{LDR}^{(d)}(l) = \frac{(\mu_{l,d}^N - \mu_{l,d}^A)^2}{(\sigma_{l,d}^N)^2 + (\sigma_{l,d}^A)^2 + \epsilon}.
    \end{equation}
    The overall class separability of layer~$l$ is the mean LDR across all~$D$ feature dimensions:
    \begin{equation}
        \text{LDR}(l) = \frac{1}{D} \sum_{d=1}^{D} \text{LDR}^{(d)}(l).
    \end{equation}
    A higher $\text{LDR}(l)$ suggests stronger linear separability between the normal and anomalous classes, implying more discriminative features at layer~$l$.

    \item \textbf{Information Concentration via Feature Entropy \cite{ali2025Entropylens}:} To assess the information concentration within the feature representations, for each feature dimension $d$ at layer $l$, we estimate the probability distribution by discretizing the feature values into a fixed number of $B$ bins with evenly spaced boundaries determined by the range of feature values across all samples. The entropy for the $d$-th dimension is then calculated as:
    \begin{equation}
        H^{(d)}(l) = -\sum_{j=1}^{B} p(\mathbf{h}_l[d] \in \text{bin}_j) \log_2 p(\mathbf{h}_l[d] \in \text{bin}_j), \tag{5}
    \end{equation}
    where $p(\mathbf{h}_l[d] \in \text{bin}_j)$ is the probability of the feature value falling into the $j$-th bin, and $\log_2$ denotes the logarithm base 2, as entropy is often measured in bits. The overall entropy for layer $l$ is the average entropy across all $D$ feature dimensions:
    \begin{equation}
        H(l) = \frac{1}{D} \sum_{d=1}^{D} H^{(d)}(l). \tag{6}
    \end{equation}
    Higher entropy values indicate more uniform distribution of feature values across bins and capturing more diverse information.

\end{itemize}

Our statistical analysis on the XD-Violence dataset revealed a consistent trend across these metrics (see Fig. \ref{fig:threemetrics}). We observed that the KL divergence, LDR, and entropy all exhibited increase in the intermediate layers of the MLLM, peaking around layer 20, and then showing a slight decrease in the deeper layers. This suggests that the statistical discriminability between normal and anomalous samples and the richness of the information captured are all maximized during intermediate layers. The subsequent decrease in deeper layers indicate that the MLLM starts to prioritize information relevant for the downstream text generation task, leading to the cost of fine-grained anomaly-related features and overall information richness for anomaly detection. More statistical results are provided in supplementary materials. These results strongly suggest a significant concentration of anomaly-relevant information within the intermediate layers of the MLLM.

\begin{figure}[t!]
    \centering
    \begin{subfigure}[b]{0.49\columnwidth}
        \includegraphics[width=\textwidth]{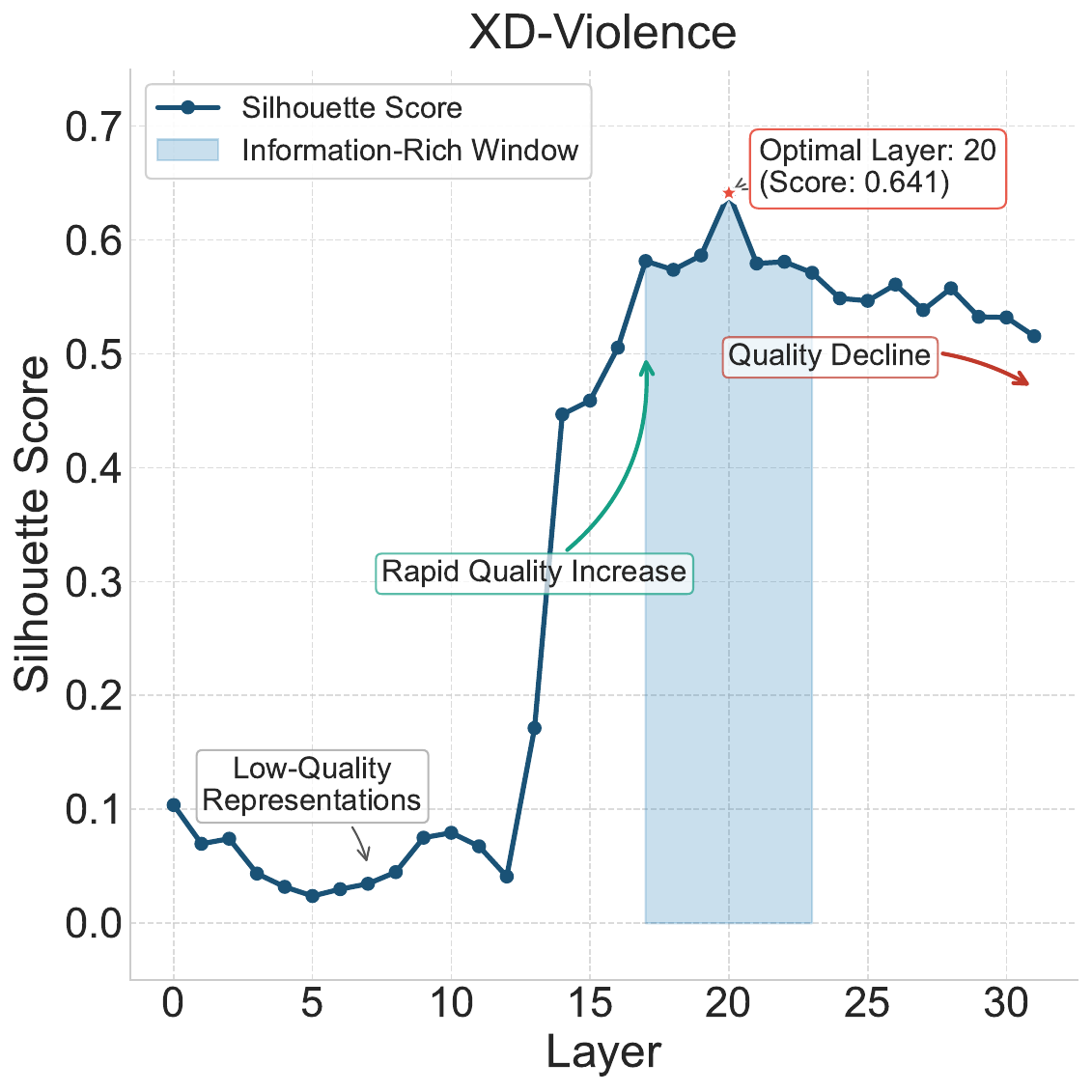}
        \label{fig:silhouette_xd}
    \end{subfigure}
    \begin{subfigure}[b]{0.49\columnwidth}
        \includegraphics[width=\textwidth]{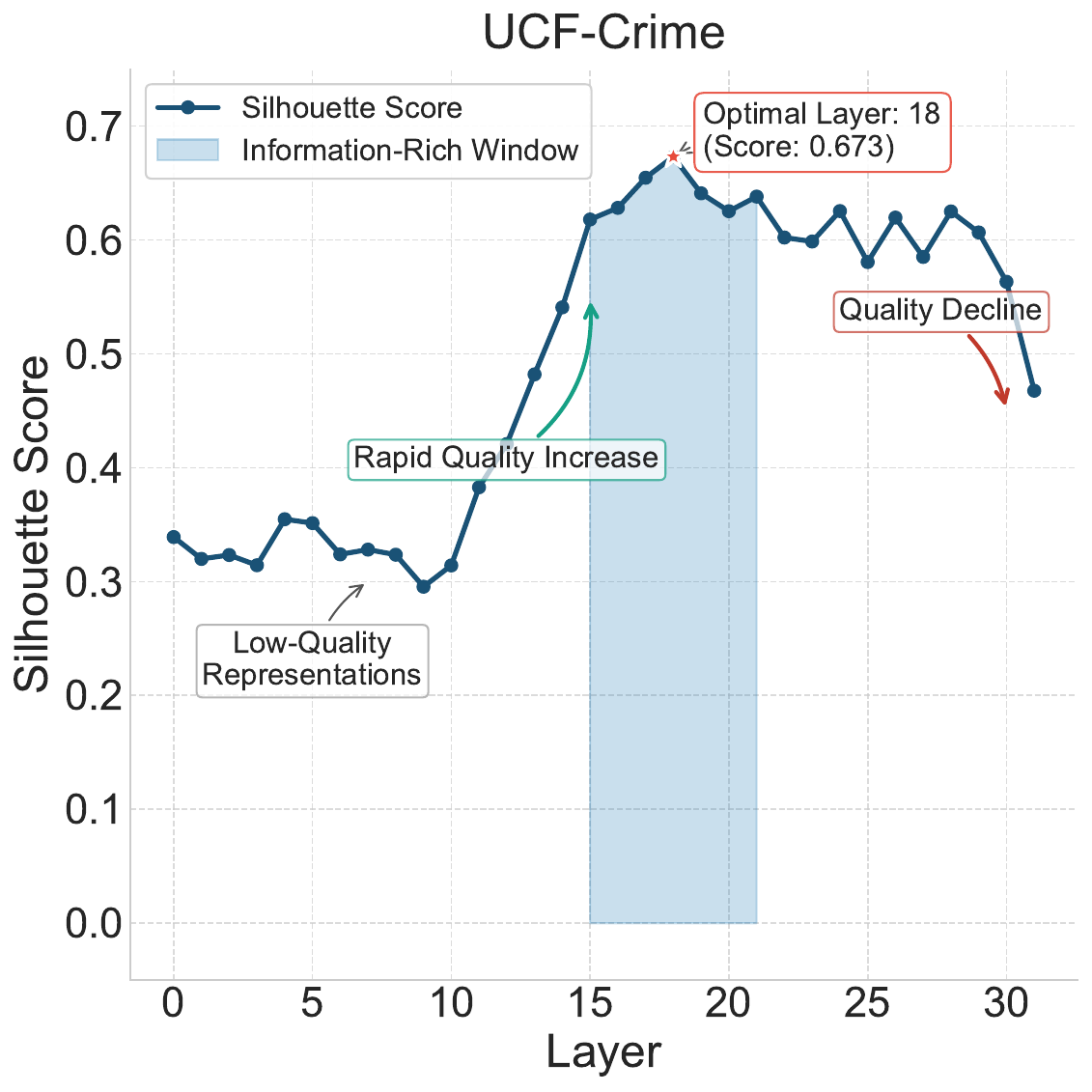}
        \label{fig:silhouette_ucf} 
    \end{subfigure}

    \caption{Silhouette score across layers on XD-Violence and UCF-Crime datasets, revealing a a trend of increasing linear separability, peaking at intermediate layer 20 before declining in deeper layers.}
    \Description{Silhouette score across layers on XD-Violence and UCF-Crime datasets, revealing a a trend of increasing linear separability, peaking at intermediate layer 20 before declining in deeper layers.}
    \label{fig:silhouette_curve} 
\end{figure}

\subsubsection{Hidden States Separability Validation}
While the statistical metrics provide quantitative evidence of layer-wise discriminability, we validate these findings from a geometric perspective by analyzing the linear separability of hidden states. $\mathbf{h}_l$. These experiments aim to provide a more intuitive results of how the normal and anomalous samples are distributed across different layers.

We assessed the linear separability using the Silhouette score. This metric quantifies how well each sample clusters with its own class compared to other classes, a higher Silhouette score indicates better-defined clusters and greater linear separability. Fig. \ref{fig:silhouette_curve} shows the Silhouette score consistently peaked around layer 20 across both the XD-Violence and UCF-Crime datasets. More validation results are provided in supplementary materials. This result strongly supports our hypothesis and aligns with our statistical analysis, indicating that the intermediate layers in MLLMs exhibit superior linear separability between normal and anomalous video segments compared to both shallower and deeper layers.

\begin{figure}[t!]
    \centering
    \begin{subfigure}{0.31\columnwidth}
        \includegraphics[width=\linewidth]{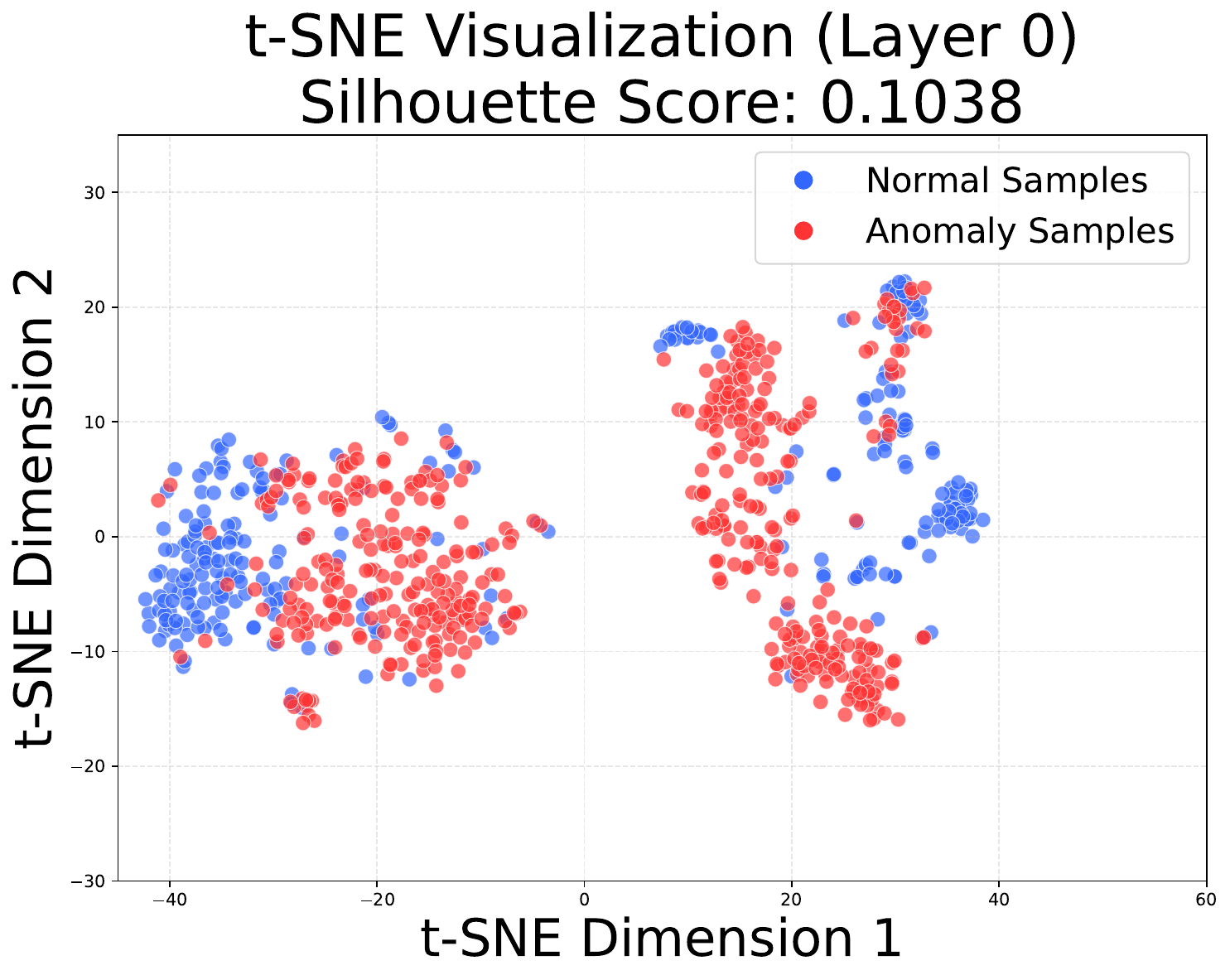}
        \label{fig:tsne0}
    \end{subfigure}
    \begin{subfigure}{0.31\columnwidth}
        \includegraphics[width=\linewidth]{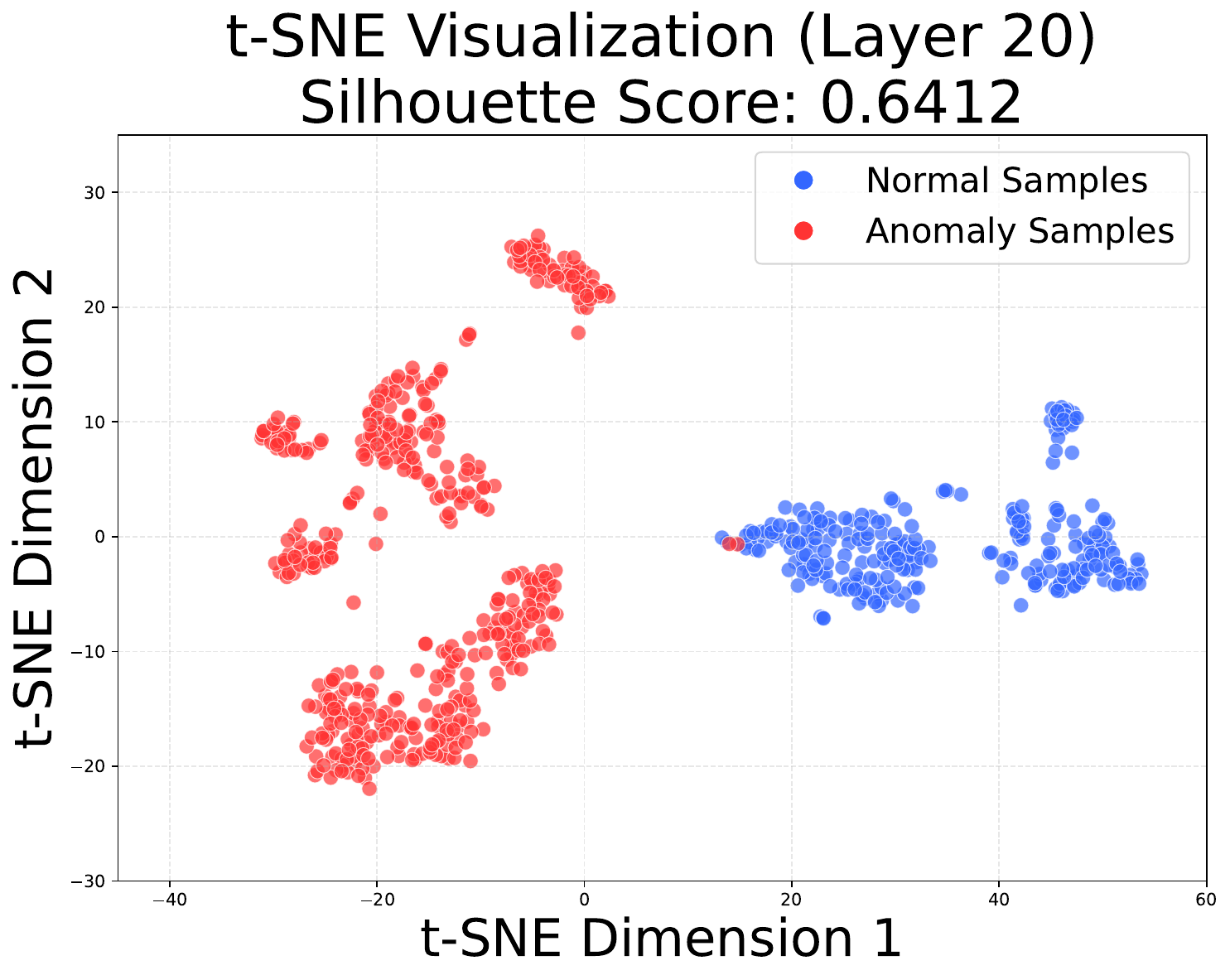}
        \label{fig:tsne20}
    \end{subfigure}
    \begin{subfigure}{0.31\columnwidth}
        \includegraphics[width=\linewidth]{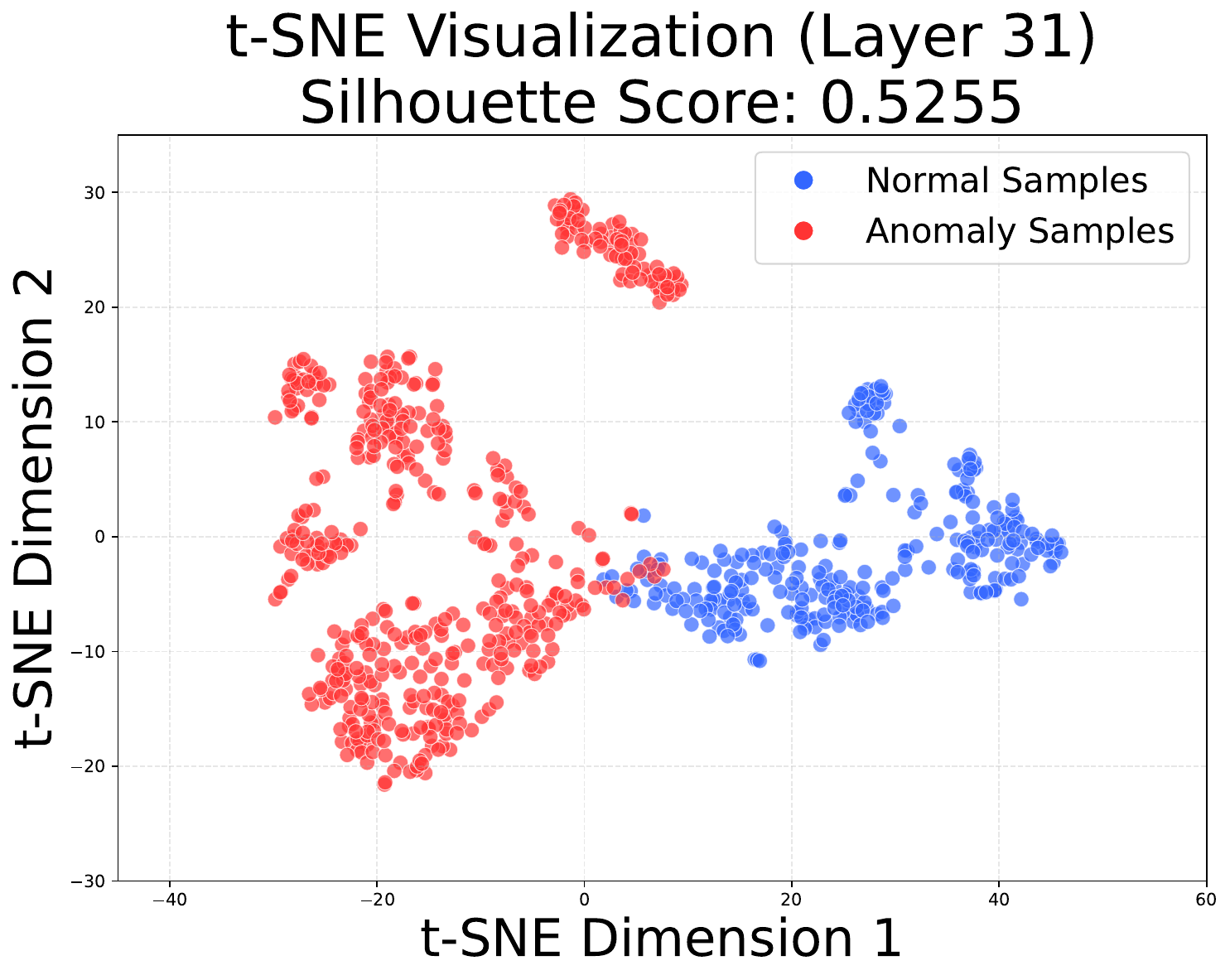}
        \label{fig:tsne31}
    \end{subfigure}
    \caption{t-SNE visualization of hidden states from the input layer (left), an optimal intermediate layer (layer 20, middle), and the output layer (right) on the XD-Violence dataset, illustrating improved separability in the intermediate layer.}
    \Description{t-SNE visualization of hidden states from the input layer (left), an optimal intermediate layer (layer 20, middle), and the output layer (right) on the XD-Violence dataset, illustrating improved separability in the intermediate layer.}
    \label{fig:tsne_visualization}
\end{figure}

We employed t-distributed Stochastic Neighbor Embedding (t-SNE) for dimensionality reduction and visualization. Fig. \ref{fig:tsne_visualization} presents the t-SNE embeddings of the hidden states extracted from the input layer (layer 0), the intermediate layer (layer 20), and the final output layer (layer 31) on the XD-Violence dataset. The visualization clearly demonstrates a progressive improvement in the separation between the clusters of normal and anomalous samples as we move from the input layer to the intermediate layer. In contrast, the feature space of the output layer shows a noticeable mixing of the two classes, suggesting a potential loss of discriminative information relevant for anomaly detection. This visual evidence effectively corroborates our quantitative findings obtained from the Silhouette score analysis, further strengthening the case for the information-rich nature of the intermediate layer representations.

\subsection{Finding: Intermediate Layer Information-rich Phenomenon in MLLMs}

Based on our empirical observations and the analysis, we solidify our finding: the \textbf{Intermediate Layer Information-rich Phenomenon}. This finding demonstrates the power and transferability of knowledge embedded within pre-trained MLLM, suggesting an inherent capability for complex tasks like anomaly detection even without task-specific fine-tuning.

\begin{figure*}[t]
    \centering
    \includegraphics[width=1\textwidth]{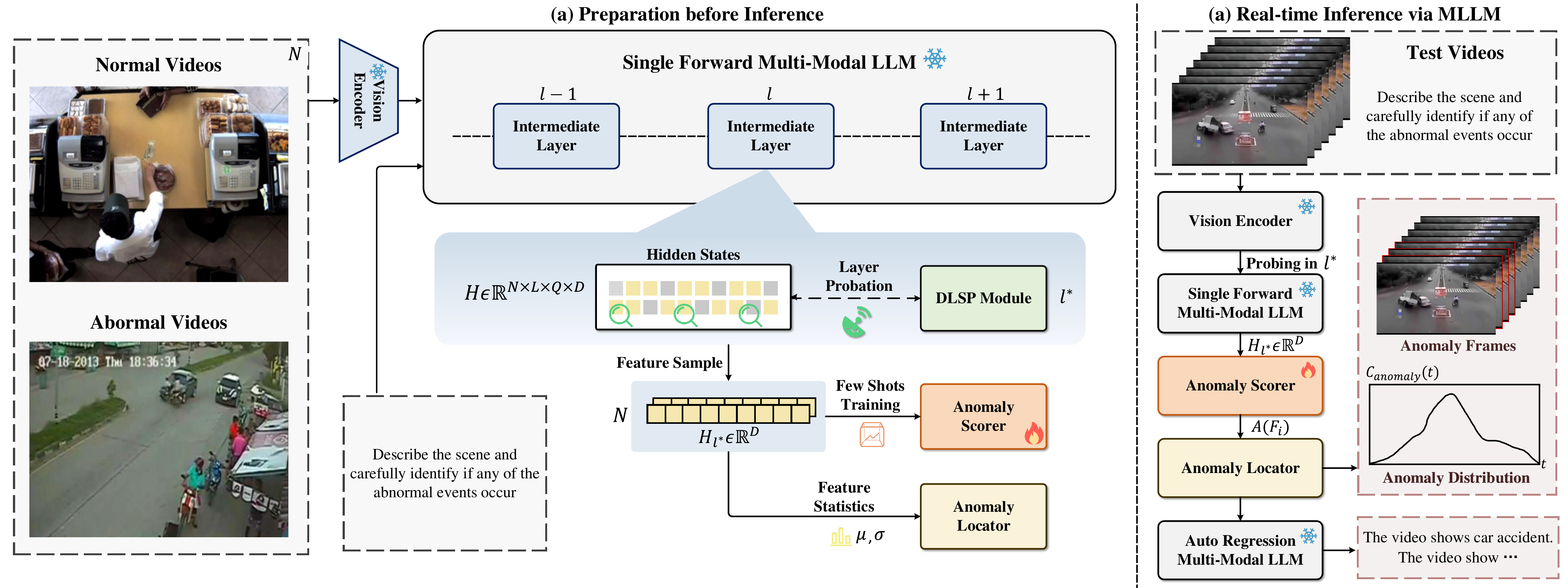} 
    \caption{Overview of the HiProbe-VAD framework. HiProbe-VAD operates in two phases: (1) Offline preparation (DLSP and scorer training) and (2) Real-time inference (frame-level scoring and localization). The DLSP module assists to input layer-wise hidden states to the scorer, which then collaborates with the temporal locator to generate final detections and descriptions.} 
    \Description{Overview of the HiProbe-VAD framework. HiProbe-VAD operates in two phases: (1) Offline preparation (DLSP and scorer training) and (2) Real-time inference (frame-level scoring and localization). The DLSP module assists to input layer-wise hidden states to the scorer, which then collaborates with the temporal locator to generate final detections and descriptions.}
    \label{fig:hiprobe}
\end{figure*}

This phenomenon can be attributed to the robust cross-modal representation learning inherent in pre-trained MLLMs. Intermediate layers appear to strike an optimal equilibrium between capturing fine-grained visual cues essential for detecting subtle anomalies and leveraging high-level semantic understanding acquired during pre-training. This balance enables these layers to effectively encode a comprehensive understanding of normative behavior, thereby retaining critical features for distinguishing deviations, while mitigating potential information loss from early fusion or over-abstraction in deeper layers optimized for text generation.

Our findings demonstrate that the intermediate layer representations of pre-trained MLLMs inherently contain sufficient information for effective video anomaly detection. This observation directly motivates the proposition of probing mechanism that leverages these information-rich intermediate hidden states, thus enabling anomaly detection without the need for computationally intensive and data-demanding fine-tuning. This core principle forms the foundational rationale for our proposed HiProbe-VAD framework.

\section{HiProbe-VAD: Tuning-Free Video Anomaly Detection via Hidden States Probing}

Building upon the Intermediate Layer Information-rich Phenomenon finding, we present HiProbe-VAD, a effective tuning-free framework for VAD leveraging pre-trained Multimodal Large Language Models (MLLMs). Fig. \ref{fig:hiprobe} illustrates the architecture of HiProbe-VAD. Our framework is designed with three key components: a Dynamic Layer Saliency Probing (DLSP) module to extract hidden states and determine the most effective intermediate layer for VAD, a Lightweight Anomaly Scorer trained with few-shot probing on the features from the selected layer to score the input frames, and a Temporal Anomaly Localization module to detect anomaly frames. Finally, we aggregates anomalous frames and subsequently generates a comprehensive description of the detected anomalies.

\subsection{Preparation with Hidden States From MLLMs}

Before the real-time inference, we need to identify the optimal layer from MLLM and train a lightweight anomaly scorer for VAD. This phase operates at the video level using the hidden states extracted from few subset of the training set to capture comprehensive information for effective layer selection and scorer training.

\subsubsection{Dynamic Layer Saliency Probing} The Dynamic Layer Saliency Probing module aims to identify the intermediate layer $l^*$ that provides the most discriminative features for distinguishing between normal and abnormal video content. This process is performed on a very few training set (about 1\%) of the training sets of UCF-Crime and XD-Violence datasets. For each video $v$ in this subset, we extract hidden states $\mathbf{H}^{(v, l)}$ at each layer $l$ during the first token generation via MLLM and then calculate the anomaly sensitivity (KL divergence), class separability (LDR), and information concentration (Entropy) of these features between normal and abnormal video samples as mentioned in Sec. \ref{metricss}.

To effectively combine these metrics, we apply Z-score normalization across all layers for KL divergence, LDR, and Entropy. For a metric $M \in \{D_{\text{KL}}(l), \text{LDR}(l), H(l)\}$, the normalized score $\text{Norm}(M(l))$ is calculated as:
\begin{equation}
    \text{Norm}(M(l)) = \frac{M(l) - \mu_M}{\sigma_M},
\end{equation}

where $\mu_M$ and $\sigma_M$ are the mean and standard deviation of the metric $M$ across all layers $\{1, \dots, L\}$. The saliency score $S(l)$ for each layer is then calculated as the sum of the normalized KL divergence, LDR and Entropy:
\begin{equation}
    S(l) = \text{Norm}(D_{\text{KL}}(l)) + \text{Norm}(\text{LDR}(l)) + \text{Norm}(H(l)).
\end{equation}

The optimal layer $l^*$ is selected to maximizes this saliency score:
\begin{equation}
    l^* = \arg\max_{l \in \{1, \dots, L\}} S(l).
\end{equation}

This video-level analysis ensures that the selected layer is robust and effective for anomaly detection across different video scenarios. The identified optimal layer index $l^*$ is then used for training the Anomaly Scorer and for real-time inference with MLLMs.

\subsubsection{Lightweight Anomaly Scorer Training} The Anomaly Scorer employs a lightweight logistic regression classifier trained offline on hidden states from optimal layer $l^*$ identified by DLSP. Let $\mathbf{h}_{l^*}^{(i)}$ denote the resampled hidden states for $i$-th sample. The predicted probability is $p_i = \sigma(\mathbf{w}^T \cdot \mathbf{h}_{l^*}^{(i)} + b)$, where $\sigma(\cdot)$ is the sigmoid function, $\mathbf{w}$ and $b$ are learned weight vector and bias. The classifier is trained to distinguish between normal ($y_i=0$) and anomalous ($y_i=1$) samples by minimizing the binary cross-entropy loss:
\begin{equation}
\mathcal{L}(\mathbf{w}, b) = -\frac{1}{N} \sum_{i=1}^{N} [y_i \log(p_i) + (1 - y_i) \log(1 - p_i)]
\end{equation}
using the LBFGS optimizer for 1000 epochs. This simple but effective model ensures the anomaly scorer is optimized to utilize the anomaly-sensitive features from the selected MLLM layer, balancing accuracy and efficiency required for real-time inference.

\subsection{Inference in HiProbe-VAD: Frame-Level Processing and Explanation}

The real-time inference phrase in MLLMs focuses on processing unseen videos to detect and localize anomaly frames, and finally provide a comprehensive anomaly description for video.

\subsubsection{Frame-Level Anomaly Scoring} For an input video, we segment it into a sequence of frames and uniformly sample keyframes from each segment. For sampled keyframes $F_i$ from each segment, we use single forward pass from MLLM to extract 
hidden states $\mathbf{h}_{l^*}(F_i)$ from the optimal layer $l^*$. The extracted features are then fed into the Lightweight Anomaly Scorer to obtain an anomaly probability $A(F_i)$ for each segment:
\begin{equation}
    A(F_i) = \sigma(\mathbf{w}^T \cdot \mathbf{h}_{l^*}(F_i) + b),
\end{equation}

where $\sigma$ is the sigmoid function, $\mathbf{w}$ and $b$ are the learned weight vector and bias of the logistic regression classifier. This frame-level scoring provides a temporal sequence of anomaly probabilities for the input video, where each frame is associated with a score indicating its probability of being anomalous.

\subsubsection{Temporal Anomaly Localization} 

To generate a comprehensive anomaly description, we aggregate the frame-level anomaly scores over time. First, we apply the Gaussian kernel smoothing to the sequence of anomaly probabilities to reduce noise and obtain a smoother anomaly probability curve $C(t)$. We then identify potential anomalous segments by applying a threshold $T$ to this smoothed curve. The threshold $T$ is determined adaptively based on the mean $\mu_A$ and standard deviation $\sigma_A$ of the anomaly scores obtained from the DLSP module on the few-shot training set:
\begin{equation}
    T = \mu_A + \kappa \cdot \sigma_A.
\end{equation}

Consecutive frames with smoothed anomaly scores above this threshold are grouped into anomalous segments. Similarly, with scores below the threshold are grouped into normal segments.

\subsubsection{Explainable VAD via MLLMs} To provide interpretable insights into the detected anomalies, we separately input anomalous segments and normal segments into auto-regression process with pre-trained MLLMs. This process transforms the video segments into precise explanations, enhancing the interpretability of the HiProbe-VAD framework and providing users with a deeper understanding of the detected abnormal activities within the video.

\section{Experiments}
\subsection{Experimental Setup}
\subsubsection{Datasets}
We evaluated our framework on two commonly used datasets for video anomaly detection: UCF-Crime \cite{sultani2018RealWorld} and XD-Violence \cite{wu2020Not}.
\begin{itemize}[noitemsep, leftmargin=*, topsep=0pt]
    \item \textbf{UCF-Crime} dataset includes 1900 untrimmed real-world surveillance videos (approximately 128 hours) with frame-level annotations, covering 13 types of anomalies. The dataset is split into 1610 training videos and 290 testing videos.
    \item \textbf{XD-Violence} includes 4754 untrimmed videos (approximately 217 hours) from movie and YouTube videos, annotated with 6 types of violent anomalies at the video level (weak labels). It consists of 3954 training videos and 800 test videos.
\end{itemize}

\subsubsection{Evaluation Metrics}
We used the Area Under the Curve(AUC) of the frame-level Receiver Operating Characteristic(ROC) as metric for the UCF-Crime dataset. For XD-Violence dataset, we used Average Precision(AP), aligning with other existing methods.

\definecolor{lightgraybg}{RGB}{240, 240, 240}
\begin{table}[t!]
\caption{Comparison of existing methods on the UCF-Crime dataset.}
\centering
\small
\setlength{\tabcolsep}{2pt}
\renewcommand{\arraystretch}{1}
\begin{tabular}{
>{\centering\arraybackslash}c
| >{\hspace{4pt}}l
@{\hspace{0pt}} c
c
}
\toprule
\textbf{Mode} & \textbf{Methods} & \textbf{Backbone} & \textbf{AUC (\%)} \\
\midrule
\multirow{10}{*}{\centering\shortstack{Weakly \\ Supervised}} 
 & Wu et al.\cite{wu2020Not} & I3D & 82.44 \\
 & MIST\cite{feng2021MIST} & I3D & 82.30 \\
 & RTFM\cite{tian2021Weaklysupervised} & I3D & 84.30 \\
 & S3R\cite{wu2022Selfsupervised} & I3D & 85.99 \\
 & MSL\cite{li2022SelfTraining} & I3D & 85.30 \\
 & UR-DMU\cite{zhou2023Dual} & I3D & 86.97 \\
 & MFGN\cite{chen2022MGFN} & I3D & 86.98 \\
 & Wu et al.\cite{wu2024Openvocabulary} & ViT & 86.40 \\
 & CLIP-TSA\cite{joo2023CLIPTSA} & ViT & 87.58 \\
 & Yang et al.\cite{yang2024Text} & ViT & 87.79 \\
 & VadCLIP\cite{wu2023VadCLIP} & ViT & 88.02 \\
\midrule
\multirow{3}{*}{\centering\shortstack{Self \\ Supervised}} 
 & TUR et al.\cite{tur2023Unsupervised} & Resnet & 66.85 \\
 & BODS\cite{wang2019GODS} & I3D & 68.26 \\
 & GODS\cite{wang2019GODS} & I3D & 70.46 \\
\midrule
\multirow{2}{*}{\centering\shortstack{Unsupervised}} 
 & GCL\cite{zaheer2022Generative} & ResNext & 71.04 \\
 & DYANNET\cite{thakare2023DyAnNet} & I3D & 84.50 \\
\midrule
\multirow{10}{*}{\centering\shortstack{Tuning-Free \\ Multimodal \\ VAD}} 
 & Zero-Shot CLIP\cite{radford2021Learning} & ViT & 53.16 \\
 & Zero-shot IMAGEBIND (VIDEO)\cite{girdhar2023ImageBind} & ViT & 55.78 \\
 & Zero-shot IMAGEBIND (IMAGE)\cite{girdhar2023ImageBind} & ViT & 53.65 \\
 & LLAVA-1.5\cite{liu2024Improved} & ViT & 72.84 \\
 & LAVAD\cite{zanella2024Harnessing} & ViT & 80.28 \\
 
 & \cellcolor{lightgraybg}\textbf{HiProbe-VAD (LLaVA-OV)\cite{li2024LLaVAOneVision}} & \cellcolor{lightgraybg}ViT & \cellcolor{lightgraybg}82.26 \\
 & \cellcolor{lightgraybg}\textbf{HiProbe-VAD (Qwen2.5-VL)\cite{bai2025Qwen25VL}} & \cellcolor{lightgraybg}ViT & \cellcolor{lightgraybg}85.89 \\ 
 & VERA\cite{ye2024VERA} & ViT & 86.55 \\
 & \cellcolor{lightgraybg}\textbf{HiProbe-VAD (InternVL2.5)\cite{chen2024Intern}} & \cellcolor{lightgraybg}ViT & \cellcolor{lightgraybg}86.72 \\

\midrule
\multirow{2}{*}{\centering\shortstack{Fine-Tuned \\ MLLM}} 
 & Holmes-VAU\cite{zhang2024HolmesVAU} & ViT & 87.68 \\
 & \cellcolor{lightgraybg}\textbf{HiProbe-VAD (Holmes-VAU)} & \cellcolor{lightgraybg}ViT & \cellcolor{lightgraybg}88.91 \\
\bottomrule
\end{tabular}
\label{tab:comparison_auc}
\end{table}

\subsubsection{Implementation Details}
We uniformly sampled $K=8$ key-frames at a fixed interval of each video segment of 24 frames. We used InternVL2.5 \cite{chen2024Intern} as the backbone MLLM for HiProbe-VAD and also conducted experiments with Qwen2.5-VL \cite{bai2025Qwen25VL}, LLaVA-OneVision \cite{liu2024Improved}, and Holmes-VAU \cite{zhang2024HolmesVAU} backbones. The lightweight logistic regression classifier was trained on the hidden states extracted from the optimal layer identified by the DLSP module. The Gaussian kernel width $\sigma$ for temporal localization was set to 0.4, and the threshold parameter $\kappa$ was set to 0.2. All experiments were performed on a server equipped with an NVIDIA 4090 GPU.

\definecolor{lightgraybg}{RGB}{240, 240, 240}

\begin{table}[t!]
\caption{Comparison of existing methods on the XD-Violence dataset.}
\centering
\small
\setlength{\tabcolsep}{2pt}
\renewcommand{\arraystretch}{1}
\begin{tabular}{
>{\centering\arraybackslash}c
| >{\hspace{4pt}}l
@{\hspace{0pt}} c
c
}
\toprule
\textbf{Mode} & \textbf{Methods} & \textbf{Backbone} & \textbf{AP (\%)} \\
\midrule
\multirow{10}{*}{\centering\shortstack{Weakly \\ Supervised}} 
 & Wu et al.\cite{wu2020Not} & I3D & 73.20 \\
 & RTFM\cite{tian2021Weaklysupervised} & I3D & 77.81 \\
 & MSL\cite{li2022SelfTraining} & I3D & 78.28 \\
 & MFGN\cite{chen2022MGFN} & I3D & 79.19 \\
 & S3R\cite{wu2022Selfsupervised} & I3D & 80.26 \\
 & UR-DMU\cite{zhou2023Dual} & I3D & 81.66 \\
 & Wu et al.\cite{wu2024Openvocabulary} & ViT & 66.53 \\ 
 & CLIP-TSA\cite{joo2023CLIPTSA} & ViT & 82.19 \\
 & Yang et al.\cite{yang2024Text} & ViT & 83.68 \\
 & VadCLIP\cite{wu2023VadCLIP} & ViT & 84.51 \\
\midrule
\multirow{8}{*}{\centering\shortstack{Tuning-Free \\ Multimodal \\ VAD}} 
 & Zero-Shot CLIP\cite{radford2021Learning} & ViT & 17.83 \\
 & Zero-shot IMAGEBIND (VIDEO) \cite{girdhar2023ImageBind}& ViT & 25.36 \\
 & Zero-shot IMAGEBIND (IMAGE) \cite{girdhar2023ImageBind}& ViT & 27.25 \\
 & LLAVA-1.5\cite{liu2024Improved} & ViT & 50.26 \\
 & LAVAD\cite{zanella2024Harnessing} & ViT & 62.01 \\
 & \cellcolor{lightgraybg}\textbf{HiProbe-VAD (LLaVA-OV)\cite{li2024LLaVAOneVision}} & \cellcolor{lightgraybg}ViT & \cellcolor{lightgraybg}76.32 \\
 & \cellcolor{lightgraybg}\textbf{HiProbe-VAD (Qwen2.5-VL)\cite{bai2025Qwen25VL}} & \cellcolor{lightgraybg}ViT & \cellcolor{lightgraybg}80.94 \\
 & \cellcolor{lightgraybg}\textbf{HiProbe-VAD (InternVL2.5)\cite{chen2024Intern}} & \cellcolor{lightgraybg}ViT & \cellcolor{lightgraybg}82.15 \\

\midrule
\multirow{2}{*}{\centering\shortstack{Fine-Tuned \\ MLLM}} 
 & Holmes-VAU \cite{zhang2024HolmesVAU}& ViT & 88.96 \\
 & \cellcolor{lightgraybg}\textbf{HiProbe-VAD (Holmes-VAU)} & \cellcolor{lightgraybg}ViT & \cellcolor{lightgraybg}89.51 \\
\bottomrule
\end{tabular}
\label{tab:comparison}
\end{table}

\subsection{Performance and Comparisons}

\subsubsection{Comparison with State-of-the-arts} Tab. \ref{tab:comparison_auc} presents a comparison of HiProbe-VAD with state-of-the-art methods on the UCF-Crime dataset. Results show that our framework outperforms all existing tuning-free methods. HiProbe-VAD using the InternVL2.5\cite{chen2024Intern} backbone achieves an AUC of 86.72\%, representing improvement of +6.44\% compared to LAVAD and a +0.17\% improvement over the VERA. Furthermore, our framework significantly outperforms all existing unsupervised and self-supervised methods. Notably, HiProbe-VAD surpasses several weakly supervised methods that rely on substantial labeled data, demonstrating its strong performance with limited data for layer selection and scorer training.

Tab. \ref{tab:comparison} shows the comparison with state-of-the-art methods on the XD-Violence dataset. Similar to the results on UCF-Crime, HiProbe-VAD exhibits competitive performance. In tuning-free methods HiProbe-VAD stands a promising pipeline across MLLMs in VAD, achieving significant performance without any fine-tuning of MLLMs and without requiring a large amount of labeled data.


\begin{table}[t!]
\caption{Zero-shot experiments on the XD-Violence and UCF-Crime datasets.}
\centering
\setlength{\tabcolsep}{4pt}
\renewcommand{\arraystretch}{1}
\begin{tabular}{
>{\centering\arraybackslash}c | 
>{\centering\arraybackslash}c 
>{\centering\arraybackslash}c
}
\toprule
\multirow{2}{*}{\centering \textbf{Method}} & \multicolumn{1}{c}{\textbf{XD-Violence}} & \multicolumn{1}{c}{\textbf{UCF-Crime}} \\
\cmidrule(l{0pt}r{0pt}){2-3}
& \textbf{AP (\%)} & \textbf{AUC (\%)} \\
\midrule
LLaVA-OneVision\cite{li2024LLaVAOneVision} & 71.17 & 76.21 \\
Qwen2.5-VL\cite{bai2025Qwen25VL} & 75.86 & 80.60 \\
InternVL2.5\cite{chen2024Intern} & 77.04 & 81.35 \\
Holmes-VAU\cite{zhang2024HolmesVAU} & 85.65 & 86.33 \\
\bottomrule
\end{tabular}
\label{tab:zero_shot}
\end{table}

\subsubsection{Cross-Model Generalization} To evaluate cross-model generalization capability, we conducted experiments using three different pre-trained MLLMs. As shown in Tab. \ref{tab:comparison_auc} and \ref{tab:comparison}, the InternVL2.5 backbone achieved the best performance. The Qwen2.5-VL-based and the LLaVA-OneVision-based\cite{li2024LLaVAOneVision} framework also show competitive results compared to existing methods. These results demonstrate the robustness and adaptability of our approach across diverse MLLM architectures without any fine-tuning with MLLMs. To further explore the adaptability of our framework, we employed the fine-tuned Holmes-VAU as the backbone, results revealed that our framework achieved state-of-the-art performance compared all existing methods, highlighting the promising potential of HiProbe-VAD as a robust and high-performing solution for video anomaly detection across various MLLM backbones.

\subsubsection{Zero-shot Generalization Capability} We further investigated the zero-shot generalization capability of HiProbe-VAD by training UCF-Crime dataset only and test on XD-Violence dataset and vice versa. Tab. \ref{tab:zero_shot} presents results that HiProbe-VAD achieves an AUC of 81.35\% on UCF-Crime and an AP of 77.04\% on XD-Violence in the zero-shot setting. Similarly, Qwen2.5-VL, LLaVA-OneVision and Holmes-VAU backbones also demonstrate promising zero-shot performance, suggesting that the intermediate hidden states of these pre-trained models inherently capture transferable anomaly-related features, enabling effective generalization to unseen datasets without task-specific adaptation and reducing the need for extensive labeled data collection in new environments.

\begin{figure}[t!]
    \centering
    \begin{subfigure}{1\columnwidth}
        \includegraphics[width=\linewidth]{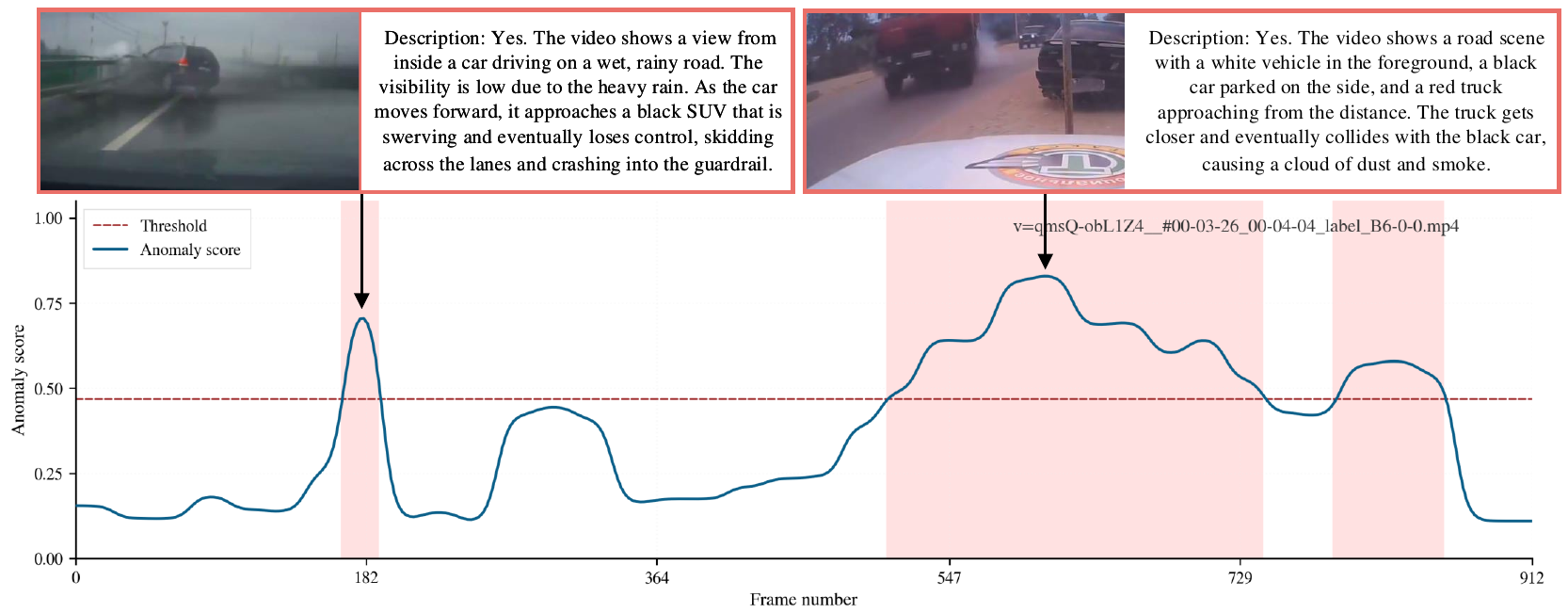}
        \label{fig:resultabnormal}
    \end{subfigure}
    \begin{subfigure}{1\columnwidth}
        \includegraphics[width=\linewidth]{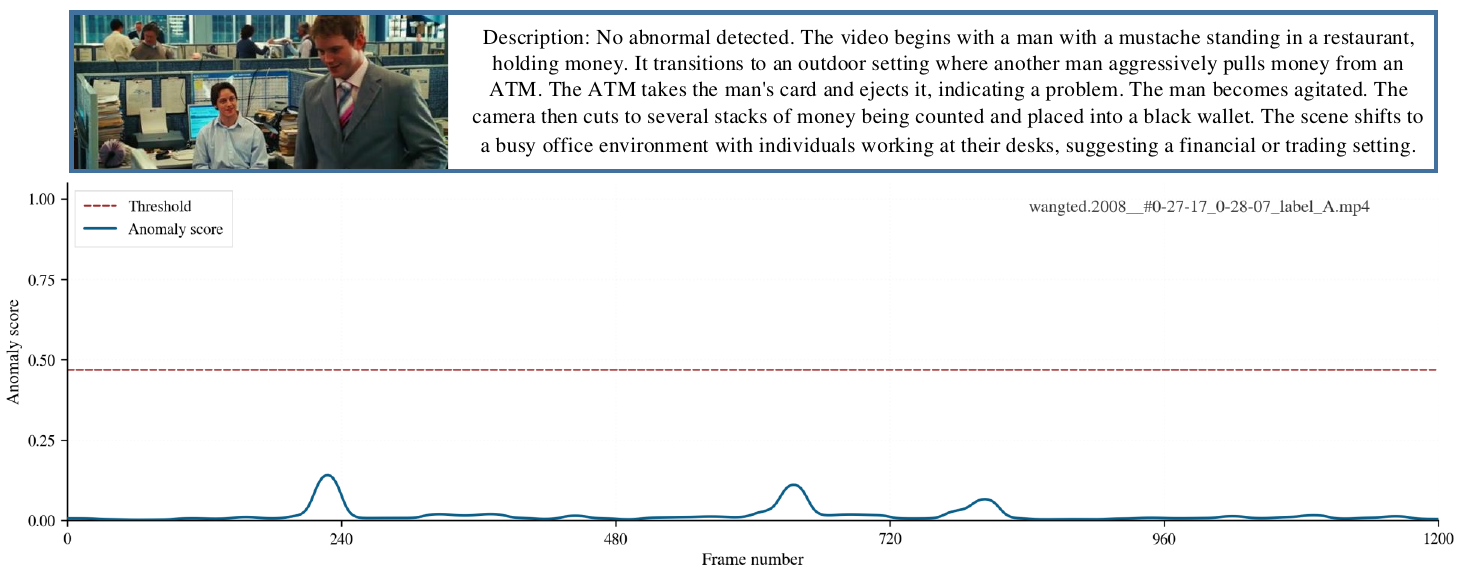}
        \label{fig:resultnormal}
    \end{subfigure}
    \caption{Qualitative results of HiProbe-VAD on XD-Violence dataset. Each panel shows a representative video snippet, the corresponding anomaly curve generated by our framework, The shaded regions in the abnormal video plot highlight the time intervals where the anomaly score exceeds the threshold, indicating detected anomalous segments. Further generated descriptions are also provided.}
    \Description{Qualitative results of HiProbe-VAD on XD-Violence dataset. Each panel shows a representative video snippet, the corresponding anomaly curve generated by our framework, The shaded regions in the abnormal video plot highlight the time intervals where the anomaly score exceeds the threshold, indicating detected anomalous segments. Further generated descriptions are also provided.}
    \label{fig:combined}

\end{figure}

\subsubsection{Qualitative Results} 
Fig. \ref{fig:combined} presents qualitative results on abnormal and normal test video from XD-Violence dataset, offering intuitive visual results to our framework. For each video, the plot shows the anomaly curves across different frames. For the abnormal video, the plot shows a fluctuating anomaly score curve, with red shaded regions indicating the detected anomalous segments where the anomaly score exceeds the learned threshold, effectively pinpointing the moments of unusual activity. The normal video exhibits consistently low anomaly scores. The corresponding descriptions generated by the MLLM are shown above, demonstrating the potential for integrating our anomaly detection framework with high-level semantic understanding of the video content. More results and analyses are provided in supplementary materials.

\subsection{Ablation Studies}

\begin{table}[t!]
\caption{Ablation Study of HiProbe-VAD on UCF-Crime and XD-Violence datasets (using InternVL2.5).}
\centering
\begin{tabular}{l|cc}
\hline
\multirow{2}{*}{Ablation Setting} & \multicolumn{1}{c}{UCF-Crime} & \multicolumn{1}{c}{XD-Violence} \\
& AUC (\%) & AP (\%) \\
\hline
\textbf{Full HiProbe-VAD} & \textbf{86.72} & \textbf{82.15} \\
\hline
\multicolumn{3}{l}{\textit{w/o. Dynamic Layer Saliency Probing (DLSP)}} \\
\hline
Fixed Last Layer & 83.21 & 79.28 \\
Fixed Mid Layer(Layer 16) & 78.60 & 75.61 \\
\hline
\multicolumn{3}{l}{\textit{w/o. Lightweight Anomaly Scorer}} \\
\hline
SVM & 84.87 & 80.63 \\
Distance-based Scoring & 80.34 & 75.65 \\
\hline
\multicolumn{3}{l}{\textit{w/o. Temporal Localization}} \\
\hline
Fixed Threshold = 0.75 & 70.42 & 65.43 \\
Fixed Threshold = 0.5 & 85.78 & 79.93 \\
Fixed Threshold = 0.25 & 76.01 & 72.45 \\
\hline
\end{tabular}
\label{tab:ablation_study}
\end{table}

To better understand the contribution of each component in our HiProbe-VAD framework, we conducted a series of ablation experiments on UCF-Crime and XD-Violence datasets using the InternVL2.5 backbone. The results are summarized in Tab.\ref{tab:ablation_study}.

\subsubsection{Effectiveness of Dynamic Layer Saliency Probing} To validate the effectiveness of our Dynamic Layer Saliency Probing module in identifying the most relevant features for anomaly detection within the MLLM, we conducted ablation experiments comparing it to fixed layer selection strategies. Table \ref{tab:ablation_study} shows that using a fixed last layer of InternVL2.5 resulted in a notable performance decrease of 3.51\% in AUC on UCF-Crime and 2.87\% in AP on XD-Violence. Fixing the layer to a mid layer (layer 16) led to more substantial drops of 8.12\% and 6.54\%, respectively. These significant performance degradations highlight the effectiveness of DLSP in dynamically identifying and leveraging these information-rich layers.

\subsubsection{Impact of the Lightweight Anomaly Scorer} To assess the effectiveness of logistic regression classifier as the anomaly scorer, we compared its performance with two alternative scoring mechanisms: a Support Vector Machine (SVM) and a distance-based scoring method. As presented in Tab.\ref{tab:ablation_study}, using an SVM resulted in a decrease of 1.85\% in AUC on UCF-Crime and 1.52\% in AP on XD-Violence compared to logistic regression classifier. The distance-based scoring method exhibited lower performance, with a drop of 6.38\% in AUC on UCF-Crime and 6.50\% in AP on XD-Violence. These results suggest that while both SVM and distance-based methods can capture some anomalous patterns, the logistic regression classifier proves more effective in distinguishing between normal and abnormal events based on the features extracted by our framework.

\subsubsection{Contribution of Temporal Localization} To evaluate the contribution of localization module, we ablated $T$ with fixed thresholds to determine anomalous frames instead of adaptive method. As shown in Tab.\ref{tab:ablation_study}, using a fixed threshold of 0.75 led to substantial drop of 16.30\% in AUC and 16.72\% in AP, indicating that a high static threshold misses many subtle anomalies. While a threshold of 0.5 achieved a relatively close performance on UCF-Crime (85.78\% AUC), it still lagged behind our full method by 0.94\%, and the performance on XD-Violence (79.93\% AP) was notably lower by 2.22\%. A lower threshold of 0.25 resulted in a significant drop in AUC (76.01\%) and AP (72.45\%), lead to an increased number of false positives. These results demonstrate the effectiveness of our adaptive temporal localization, which dynamically groups anomalous frames and suppresses false alarms, yielding more accurate and robust detection than fixed thresholds.

\begin{table}[t!]
\caption{Impact of Sampling Rate on HiProbe-VAD Performance with UCF-Crime dataset.}
\centering
\begin{tabular}{c|c}
\hline
Sampling Rate (K) & AUC (\%) \\
\hline
K = 2  & 76.20 \\
K = 4  & 82.51 \\
K = 8  & \textbf{86.72} \\
K = 16 & 87.01 \\
\hline
\end{tabular}
\label{tab:sampling_rate_study}
\end{table}

\subsubsection{Impact of Keyframe Sampling Rate} To investigate the influence of the number of sampled keyframes on the performance of HiProbe-VAD, we experimented with different sampling rates by varying the number of keyframes ($K$) extracted from each 24-frame video segment. As shown in Tab.\ref{tab:sampling_rate_study}, increasing the number of keyframes generally leads to improved performance, indicates that more keyframes capture richer temporal information within each segment. However, increasing the sampling rate to $K=16$ resulted in only a marginal performance gain to 87.01\%, suggesting a potential saturation point where the benefit of additional keyframes diminishes. Considering the observed trend of limited performance improvement beyond $K=8$ alongside the substantial increase in computational resources required for processing more keyframes, we opted for $K=8$ as the default setting for our experiments. This choice offers a strong balance between achieving high anomaly detection accuracy and maintaining computational efficiency.

\section{Conclusion}
In this paper, we introduced HiProbe-VAD, a novel tuning-free framework for video anomaly detection inspired by our finding of "Intermediate Layer Information-rich Phenomenon" within pre-trained MLLMs. Our framework leverages Dynamic Layer Saliency Probing module to identify optimal intermediate layer, coupled with lightweight anomaly scorer and localization module to identify anomalies and finally generate descriptions. Experiments demonstrate that HiProbe-VAD achieves state-of-the-art performance among tuning-free methods, outperforming existing unsupervised and self-supervised approaches. The remarkable cross-model generalization capability of HiProbe-VAD across diverse MLLM architectures underscores its robustness and adaptability. We hope this work inspires further exploration of intermediate MLLM representations and video anomaly detection for broader applications.


\bibliographystyle{acm}
\balance
\bibliography{hiprobevad.bbl}

\end{document}